\documentclass[letterpaper, 10 pt, conference]{ieeeconf}  % Comment this line out if you need a4paper

\IEEEoverridecommandlockouts                              
\usepackage{ifpdf}
\usepackage{cite}
\usepackage[pdftex]{graphicx}
\usepackage{amsmath}
\usepackage{amssymb}
\usepackage{algorithm}
\usepackage[noend]{algpseudocode}
\usepackage{array}
\usepackage{url}
\usepackage{multicol}
\usepackage{booktabs}
\usepackage{siunitx}
\usepackage{color}
\title{\LARGE \bf Dynamic Compressed Sensing of Unsteady Flows with a Mobile Robot}

\author{{Sachin~Shriwastav, Gregory~Snyder and Zhuoyuan~Song}% 
\thanks{*This work was supported by the U.S.~National Science Foundation under awards National Robotics Initiative 2.0 (IIS-2024928), EPSCoR Research Infrastructure (OIA-2032522), and AI Institute in Dynamic Systems (CBET-2112085).}
\thanks{S.~Shriwastav, G.~Snyder and Z.~Song are with the Department of Mechanical Engineering, University of Hawai`i at M\={a}noa, Honolulu, HI, 96822 USA. S.~Shriwastav is also with the East-West Center Foundation Scholarship program. Emails: \{{\tt\small sachins,snyderg,zsong}\}  {\tt\small @hawaii.edu}}%
}

\begin{document}

\maketitle
\thispagestyle{plain}
\pagestyle{plain}

%%%%%%%%%%%%%%%%%%%%%%%%%%%%%%%%    ABSTRACT     %%%%%%%%%%%%%%%%%%%%%%%%%%%%%%%%%%%%%
\begin{abstract}
Large-scale environmental sensing with a finite number of mobile sensors is a challenging task that requires a lot of resources and time. This is especially true when features in the environment are spatiotemporally changing with unknown or partially known dynamics. Fortunately, these dynamic features often evolve in a low-dimensional space, making it possible to capture their dynamics sufficiently well with only one or several properly planned mobile sensors. This paper investigates the problem of dynamic compressed sensing of an unsteady flow field, which takes advantage of the inherently low dimensionality of the underlying flow dynamics to reduce number of waypoints for a mobile sensing robot. The optimal sensing waypoints are identified by an iterative compressed sensing algorithm that optimizes the flow reconstruction based on the proper orthogonal decomposition modes. An optimal sampling trajectory is then found to traverse these waypoints while minimizing the energy consumption, time, and flow reconstruction error. Simulation results in an unsteady double gyre flow field is presented to demonstrate the efficacy of the proposed algorithms. Experimental results with an indoor quadcopter are presented to show the feasibility of the resulting trajectory. 

\end{abstract}

%%%%%%%%%%%%%%%%%%%%%%%%%%%%%%%    INTRODUCTION     %%%%%%%%%%%%%%%%%%%%%%%%%%%%%%%%%%%
\section{INTRODUCTION}
Environmental sensing is an important problem with a wide range of applications ranging from long-term observations like climate change monitoring and agricultural automation to short-term use cases such as disaster management and civil surveillance. The ability to generate and interpret data of a large environment promptly using very few sensors would save time and resources in many real-life applications. For applications such as marine observatories~\cite{FiorelliE:06a, PaleyDA:08a, MaK:17a} and disaster information gathering, time and operational cost are critical considerations. It is necessary to be able to understand the environment properties efficiently with limited resources.

Large-scale environmental sensing often lead to high demands for data collection due to the large dimensionalities of the target signals. Fortunately, most meaningful signals, such as images and audio, are inherently very sparse (containing mostly zeros or value small values) when transformed in another domain (For example, Fourier, wavelet, discrete cosine transform). Most of those ``zeros" can be ruled out, which means that a very small portion of the original data is retained containing the information of the entire signal. Compressed sensing avoids the unnecessary measurements by taking a very small sample of randomized measurements and reconstructing the overall information efficiently. The primary performance metric is the reconstruction efficiency, which measures how well the reconstructed signal matches the original. 

Most related research in this relatively young field with vast applications builds on the pioneer works of Donoho~\cite{donoho2006compressed}, Candes et al.~\cite{candes2006robust, candes2008an}, Baraniuk~\cite{baraniuk2007compressive}, or Berkooz et al.~\cite{berkooz1993proper}. Clark et al.~\cite{clark2018greedy} presented a pivoted QR-based greedy algorithm to place sensors for reconstruction with a cost constraint on sensor locations. Manohar et al.~\cite{manohar2019optimized} presented an extension of the multi-resolution dynamic mode decomposition (mrDMD) as a principled framework for understanding the optimal placement of sensors in complex spatial environments and/or networked configurations. Sparse representation in a library of examples with historical flow-field data with tools like proper orthogonal decomposition (POD) was used to develop methods for flow-field reconstruction in~\cite{callaham2019robust, Bai2015low, alonso2004optimal}. Collaborative and compressed mobile sensing algorithms for distributed robotic networks to perform spatial sampling of a environmental field and build a map were presented in ~\cite{leonard2007collective, nguyen2018collaborative, hummel2011mission, mostofi2009compressive}. Brunton et al.~\cite{brunton2016sparse} presented an approach to classify between data classes using optimal sensors designed from substantially subsampled data, which informs the feature-based classification. Salam and Hsieh~\cite{salam2019adaptive} presented an approach leveraging the dynamics to determine the adaptive sampling and estimation process using mobile robots. Zhang and Vorobeychik~\cite{zhang2016submodular} proposed a generalized cost-benefit (GCB) greedy mobile robotic sensing algorithm for making select sensor measurements to make predictions about an infeasible location in the environment. Weiss~\cite{weiss2019tutorial} discussed the mathematical framework, discussion and examples of POD and its use in fluid dynamics and aerodynamics. In this work, we use POD modes to compute the reconstruction error for the subset sensing locations and the mobile sensor trajectories.

\begin{figure}[t]
    \centering
    \includegraphics[width=0.7\linewidth]{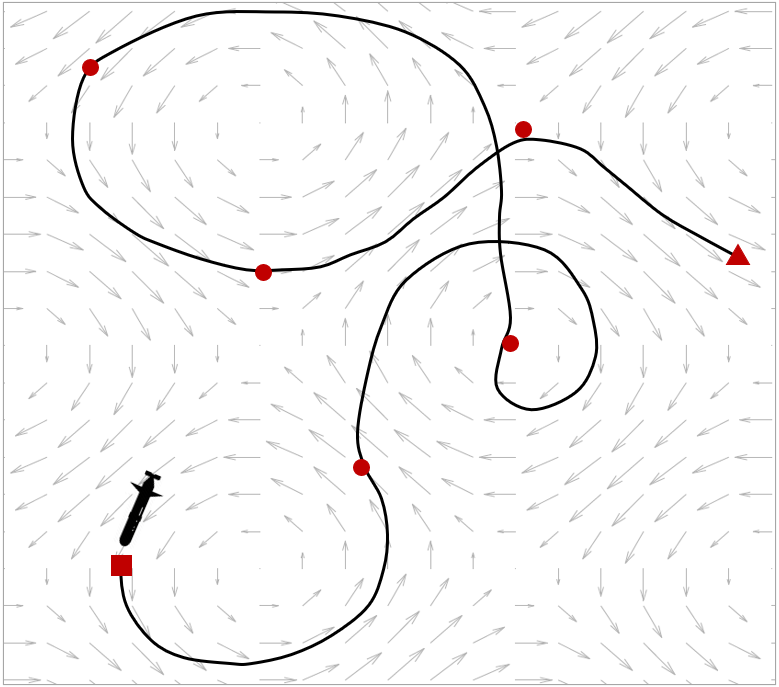}
    \caption{Illustration of dynamic compressed sensing for flow mapping with an autonomous underwater vehicle. The dots represent the optimal sensing locations determined by the compressed sensing algorithm. To collect the flow information, an optimal trajectory with the shown starting location (square) and the destination (triangle) will be chosen for the vehicle to traverse through the optimal locations as waypoints.}
    \label{fig:illustration}
\end{figure}

This paper presents the concept of dynamic compressed sensing (DCS), which is motivated by the question: ``Why do we need to place static sensors if we do not need persistent coverage of the sensing locations?". DCS is based on the hypothesis that dynamic measurements made at the right location and time are sufficient for flow reconstruction. Thus, instead of deploying static sensors at those locations, even fewer numbers of robots can be deployed to visit the optimal sensing waypoints along an optimal trajectory to collect the sensing data. Furthermore, these systems can be used to explore and identify environmental dynamics with little or no prior knowledge about the environment. An example application scenario is shown in Fig.~\ref{fig:illustration}, where an autonomous underwater vehicle is deployed to traverse through a set of optimal sensing waypoints (marked by dots) for efficient data collection. The trajectory can be optimized for specific objectives such as the minimization of time, actuation energy, or reconstruction error. 

%%%----------------- Figure:  Illustration 

Trajectory generation and optimization in strong background flows is another well-researched topic. For instance, Inanc et al.~\cite{inanc2005optimal} presented an approach for optimal path planning of autonomous underwater vehicles by studying the global flow geometry via dynamical systems methods. Senatore et al.~\cite{Senatore2008FuelefficientNI} presented an algorithm for a vehicle to track a Lagrangian coherence structure (LCS), considering this a time-dependent boundary following problem. Lolla et al.~\cite{lolla2014time} proposed a partial differential equation based methodology that predicts the time-optimal paths of autonomous vehicles in a dynamic flow. Hsieh et al.~\cite{hsieh2014distributed} presented a navigation strategy for a team of mobile sensing agents in a stochastic geophysical fluid environment using nonlinear dynamical systems tools and the LCS. Ataei and Yousefi-Koma~\cite{ataei2015three} presented an algorithm that generates three-dimensional optimal paths offline for waypoint guidance of a miniature AUV.

The proposed approach extends and combines compressed sensing and trajectory optimization to formulate a solution to large-scale flow sensing problems. We focus on demonstrating the feasibility of DCS and assume prior knowledge about the flow field is available to allow the extraction of its POD modes. An iterative compressed sensing algorithm identifies a small set of optimal sensing locations by minimizing the flow reconstruction error using only flow observation at these sensing locations. A trajectory optimization algorithm then uses these sensing locations as waypoints to generate an optimal trajectory for a mobile sensor. The objective is to further minimize the reconstruction error by taking measurements along the entire trajectory, the overall energy cost of the trajectory, and the total time along the trajectory. The major contributions of this work are: (i) Dynamic compressed sensing of a flow field using mobile robot, (ii) Optimal trajectory planning based on POD based reconstruction, (iii) Use of a mobile sensor to perform compressed sensing instead of multiple static sensors to reconstruct a time-varying unsteady flow field, (iv) emulated experimental demonstration using a indoor robotic system. 

% The remainder of the paper is organized as follows. Section~\ref{sect:preliminary} presents the preliminaries of compressed sensing and POD-based reconstruction. Section~\ref{sect:methodology} introduces the method of DCS and trajectory optimization for DCS-based flow sensing. Section~\ref{sect:simulation} presents the simulation results and analysis in an unsteady double gyre flow field.  Section~\ref{sect:experiment} demonstrates the feasibility of the optimal trajectory found in Section~\ref{sect:simulation} using a quadcopter. Finally, Section~\ref{sect:conclusion} concludes the paper with a discussion over future research directions.

%%%%%%%%%%%%%%%%%%%%%%%%%%%%    PRELIMINARIES    %%%%%%%%%%%%%%%%%%%%%%%%%%%%%%%%%%%%%%%
\section{PRELIMINARIES} \label{sect:preliminary}
The proposed DCS method for flow field sensing consists of two stages: (i) waypoints generation using compressed sensing, and (ii) trajectory optimization for POD-based flow reconstruction. In this section, we provide a brief overview of the fundamentals about compressed sensing and POD. 

%%%%%%%%%%%%%%====================  Compressed Sensing
\subsection{Compressed Sensing}
Compressed sensing is a method to reduce the number of sensors compared to a full-scale sensor deployment, while still efficiently reconstructing a signal. Compressed sensing theory asserts that, under certain conditions, one can recover the signals with far fewer samples than required by the Nyquist–Shannon sampling theorem~\cite{donoho2006compressed}. Mathematically, compressed sensing finds the coefficients of a large sparse vector which is consistent with the highly-reduced, randomly-sampled measurements and then use the equivalency of that sparse vector to invert it to recover the original signal using an appropriate transform basis. This approach is primarily based on the assumption that the signals under consideration are inherently sparse and compressible, which is almost always the case for most natural signals.

Let us consider the case where data from a large set of $n$ possible measuring locations is represented by the data vector $\mathbf{x} \in \mathbb{R}^{n}$. This data vector can be represented in an alternate transform domain using a universal transform basis $\Psi \in \mathbb{R}^{n \times n}$ and a vector $\mathbf{s}$ as 
\begin{equation} \label{eq-comp-sens-def}
    \mathbf{x} = \Psi \mathbf{s}.
\end{equation}
Vector $\mathbf{s} \in \mathbb{R}^{n}$ is $k$-sparse since it contains $k$ ($k\ll n$) non-zero elements for the data in $\mathbf{x}$ in the transformed domain. The data representations $\mathbf{x}$ and $\mathbf{s}$ are equivalent and can be transformed back and forth using $\Psi$. The objective of compressed sensing is to find $\mathbf{x}$ (full-sized signal) using a highly reduced measurement subset $\mathbf{y} \in \mathbb{R}^{m} \text{, where } k < m \ll n$, such that
\begin{equation} \label{eq-comp-sens-def2}
    \mathbf{y} = C \mathbf{x} =  C \Psi \mathbf{s} = \Theta \mathbf{s}.
\end{equation}
Here $C \in \mathbb{R}^{m \times n}$ is the measurement matrix, which essentially acts as a data mask for $\Psi \mathbf{s}$ to excite its columns and generate the sparse vector $\mathbf{s}$ that is most consistent with the measurement vector $\mathbf{y}$. The matrix $C$ is usually chosen to be a random matrix (for example random delta impulses, random Gaussian, Bernoulli) to excite most columns of $\Psi$ for better reconstruction. For efficient reconstruction, the number of measurements has to be higher than $k$, which is given by $m \propto \mathcal{O}(k log(n/k))$. Also, since $k < m \ll n$, the problem is under-determined, and there are infinitely many possible solutions for $\mathbf{s}$ that would be consistent with $\mathbf{y}$. The objective is to obtain the sparsest $\mathbf{s}$ through the following optimization problem
\begin{equation} \label{eq-comp-sens-opt1}
    \hat{\mathbf{s}} = \underset{\mathbf{s}}{\mathrm{argmin}} \ ||C \Psi \mathbf{s}||_2 + \gamma ||\mathbf{s}||_1 , 
\end{equation}
where $\gamma$ is a weighting constant for the added penalty term. This can be recast into 
\begin{equation} \label{eq-comp-sens-opt2}
    \hat{\mathbf{s}} = \underset{\mathbf{s}}{\mathrm{argmin}} \ ||\mathbf{s}||_1 \quad \textit{ s.t. } \quad ||C \Psi \mathbf{s} - \mathbf{y}||_2 = 0. 
\end{equation}
In the case where measurement noise ($\epsilon$) is present, the optimization constraint becomes 
$||C \Psi \mathbf{s} - \mathbf{y}||_2 < \epsilon$. It is to be noted that the optimization problem intends to minimize the $L_1$-norm of $\mathbf{s}$, instead of the $L_2$-norm.

Although the $L_2$-norm can generate the lowest reconstruction error, it keeps all the coefficient in $\mathbf{s}$ active by spreading the error, which is not desirable for sparsity. On the other hand, $L_0$-norm simply counts the non-zero coefficients by definition, but it makes the resulting optimization problem non-convex. Consequently, the $L_1$-norm is used, as it tends to intersect the solution space of the under-determined problems like this at sparse solutions, and hence promotes sparsity of $\mathbf{s}$, which is crucial for the optimality of the subsampled sensing location set.

%%%%%%%%%%%%%%====================  POD Modes
\subsection{Proper Orthogonal Decomposition (POD)}
We reconstruct the flow field and quantify the reconstruction error using the proper orthogonal decomposition (POD) modes, which can be obtained from existing flow models or historical data of the target area assumed to be known in this work. One of the major objectives of POD is to decompose the vector field representing turbulent fluid motion into a set of deterministic functions (POD modes) with different time and length scales that each captures portion of the total kinetic energy of the flow, with a goal to provide some inside about the organization and composition of the flow \cite{weiss2019tutorial}. 

Let us denote the 2D position by $\mathbf{x} = (x,y)$, velocity by $\mathbf{u} = (u,v)$, and time by $t$. For a flow velocity vector field $\mathbf{u}'(\mathbf{x},t)$, where $\mathbf{u}' \in \mathbb{R}^n$ is the centered flow velocity given by $\mathbf{u}$ minus its temporal mean $\Bar{\mathbf{u}}$, the proper orthogonal decomposition (with $r \leq n$ modes) is performed as
\begin{equation} \label{eq-pod-defn}
    \mathbf{u}'(\mathbf{x},t) \approx \sum_{k=1}^{r} a_k(t) \phi_k(\mathbf{x}),
\end{equation} 
where $\phi_k(\mathbf{x})$ are the POD modes and $a_k(t)$ are their modulating time coefficients. These modes can be obtained by performing singular value decomposition (SVD) of the data matrix $X \in \mathbb{R}^{m \times n}$. Alternately, it can be obtained by eigenvalue decomposition of the covariance of the data matrix. The covariance $X_c$ is obtained as
\begin{align} \label{eq-pod-cov}
    & X_c = \dfrac{1}{n-1}XX^T,
\end{align}
and the eigen-decomposition of the covariance matrix can be found as
\begin{align} \label{eq-pod-cov-eigen}
    & X_c V = V \Lambda,
\end{align}
where $\Lambda \in \mathbb{R}^{m \times m}$ is a diagonal matrix with the eigenvalues $\lambda_i$ as the diagonal terms and $V \in \mathbb{R}^{m \times m}$ are the vector of eigenvalues and column matrix of eigenvectors, respectively. The POD modes can be calculated as $\lambda_i\mathbf{v_i}$, where $\lambda_i$ and $\mathbf{v_i}$ are the $i$-th eigenvalue and eigenvector, respectively. As the data is sparse, very few of these modes constitute most of the flow field's total energy, hence most of the information of the data matrix. The temporal behavior of the modes can be obtained by
\begin{equation} \label{eq-pod-temporal-behavior}
    M_T = V^T X,
\end{equation}
which can be used to check the energy level of each mode at a given time instant while considering a sensing location's contribution to reconstruction. Each location contributes to the energy of each mode in some capacity at every instant, and that is what provides for the reconstruction efficiency. For a time series data matrix $X$, where each row contains the full time series data for at a given location, the coefficients of POD modes are time-dependent, reflecting the time-dependency of unsteady flow fields. These POD modes are natural coordinates inherent to the flow field that allow for sparse representations and reconstruction. These modes will also be used to calculate the reconstruction errors in Section~\ref{sect:methodology}-A and Section~\ref{sect:methodology}-B.

%%%%%%%%%%%%%%%%%%%%%%%%%%%%%%%    METHODOLOGY     %%%%%%%%%%%%%%%%%%%%%%%%%%%%%%%%%%%
\section{METHODOLOGY} \label{sect:methodology}
The proposed DCS-based flow sensing algorithm can be divided into two stages. In the first stage, the user defines a compressed sensing algorithm selects an optimal set of sensing locations over the area while minimizing the reconstruction error using only measurements at these locations. The total number of sensing location can be prescribed or automatically decided by the optimization through thresholding. In the second stage, a trajectory generation and optimization algorithm then generates an optimal trajectory passing through those sensing locations as waypoints while minimizing the total actuation energy, the total travel time, and flow reconstruction error, as discussed below, and summarized in Algorithm~\ref{alg1}.

%%%%%%%%%%%%%%====================  Dynamic Compressed Sensing
\subsection{Dynamic Compressed Sensing}
The selection of optimal sensing locations using compressed sensing is governed by \eqref{eq-comp-sens-def2} and \eqref{eq-comp-sens-opt2}, as discussed in Section~\ref{sect:preliminary}-A. To compute the transform matrix $\Psi$, the SVD of the reference data matrix $X \in \mathbb{R}^{m \times n}$ is computed and the left singular matrix is used as $\Psi$, that is,
\begin{align} \label{eq-svd-def}
    X = U \Sigma V^T,
\end{align}
and $\Psi = U$.
Each column of $X$ represents a time snapshot of the entire flow field. The $\Psi$ matrix is used with a randomized measurement matrix $C \in \mathbb{R}^{m \times n}$, where $n$ and $m$ are the original and down-sampled data dimensions, respectively, and $m\ll n$. The optimal set of subsampled sensing locations, $\hat{\mathbf{s}}$, can be found from \eqref{eq-comp-sens-opt2}. The sparse vector $\hat{\mathbf{s}}$ is then sorted, and the locations corresponding to the $m$ largest entries in $\hat{\mathbf{s}}$. Since a randomized measurement matrix is used, $\hat{\mathbf{s}}$ is only optimized for the an instance of the measurement matrix. An iterative step is added to further optimize among multiple $\hat{\mathbf{s}}$ solutions obtained from different random instances of measurement matrix based on based on the reconstruction performance. The algorithm tests a prescribed number of location subsets and calculates the reconstruction error for each one, and finally chooses the one with best reconstruction performance. 
For reconstruction using $N_1$ sensor locations, the approximated measurement at a sensing location $\mathbf{x}_i, \ i \in [1,2, \ldots, N_1]$ using $N_2$ POD modes $\phi_k, \ k \in [1,2, \ldots, N_2]$ is given by
\begin{equation} \label{eq-SenLoc-approximation}
    \hat{\mathbf{u}}_i = \sum_{k=1}^{N_2} \ a_k(t) \phi_k(\mathbf{x}_i).
\end{equation}
The net flow reconstruction error for the sensor subset, 
\begin{equation} \label{eq-SenLoc-RecErr}
    \mathbf{e}_{\mathbf{x}} = |\hat{\mathbf{u}} -\mathbf{u}^*|,
\end{equation}
can then be minimized, where $\mathbf{u}^*$ is the true flow velocity, which is used in \eqref{eq-SenLoc-approximation} to calculate the POD mode coefficients ($a_k(t)$). The approximated flow velocity values for the grid ($\hat{\mathbf{u}}$) are then calculated using $a_k(t)$ and the corresponding POD modes \eqref{eq-SenLoc-approximation}.

Once the reconstruction errors for all the trials are obtained, the waypoint set corresponding to the smallest reconstruction error is chosen as the optimal solution ($\mathbf{y}^*$)  for most effective reconstruction. The core idea of dynamic compressed sensing is to define an optimal trajectory for the mobile robot through these locations as waypoints, instead of placing static sensors. The robot takes measurements along the trajectory, which is then used for reconstruction of the entire environment.

%%%%%%%%%%%%%%====================  Trajectory Optimization
\subsection{Trajectory Optimization}
\begin{figure}[t]
    \centering
    \includegraphics[width=0.95\linewidth]{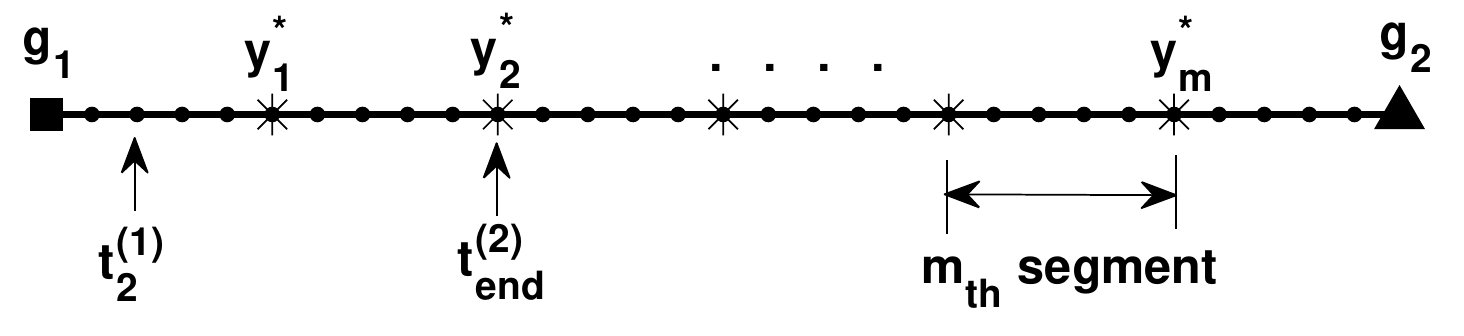}
    \caption{Illustration of a sample trajectory set up for optimization. The square, triangle, star and the dot markers denote the starting location ($\mathbf{g}_1$), the goal location ($\mathbf{g}_2$), the optimal waypoints to be visited in a prescribed sequence ($\mathbf{y}^*_k, \ k \in {1,2,\ldots,m}$) and the trajectory time instants ($\mathbf{z}(t_i)$), respectively. The trajectory (including $\mathbf{g}_1$ and $\mathbf{g}_2$) with total T time steps is divided into $(m+1)$ segments of equal time instants. The $i$-th time step of the $k$-th segment is denoted as $t_i^{(k)}$.}
    \label{fig:traj_illustration}
\end{figure}

Trajectory planning takes into account the minimization of the flow reconstruction error, energy cost, and time, while enforcing the robot to traverse through all the designated waypoints along a smooth trajectory. The trajectory is constrained to be a piece-wise continuous spline. The state vector of the optimization problem is defined as $\mathbf{p} = [\mathbf{x}^T,t]^T$, such that, $\mathbf{p}(t_i) = [x(t_i), y(t_i)]^T$. The set of $m$ optimal sensing locations (waypoints) is defined as $\mathbf{y}^* = [\mathbf{p}_1^T, \mathbf{p}_2^T, \ \ldots, \ \mathbf{p}_m^T]^T$. Equality constraints are used to enforce the trajectory passing through all sensing locations, and its setup has been shown in Fig.~\ref{fig:traj_illustration}. The prescribed starting and goal locations are denoted by $\mathbf{g}_1$ and $\mathbf{g}_2$, respectively, such that, $\mathbf{g} = (x,y)^T$. After the segments between the consecutive waypoints are defined, the last time step of each segment of the trajectory $\mathbf{z}$ has to coincide with the corresponding waypoint, that is, $\mathbf{z}(t^{(k)}_\text{end}) = \mathbf{y}^*_k, \ k \in {1,2, \ldots , m}$, where $t_i^{(k)}$ is the $i$-th time step of the $k$-th segment. Similarly, the first and last time steps of the trajectory has to coincide with $\mathbf{g}_1$ and $\mathbf{g}_2$, respectively.

The flow velocity are measured along the trajectory to perform flow field reconstruction. The cost function of trajectory optimization consists of three penalty terms minimizing the reconstruction error ($E$), actuation energy ($F$), and the total time ($D$). 

For a candidate trajectory $\mathbf{z}$ with a total of $T$ time steps, at each time step $t_i$, $i \in [0,\ldots, T]$, the flow reconstruction error is calculated as
\begin{equation} \label{eq-traj-err1}
    \mathbf{e}(t_i) = |\hat{\mathbf{u}}(\mathbf{x}(t_i), t_i) -\mathbf{u}^*(\mathbf{x}(t_i), t_i)|.
\end{equation}
This is done in following steps: (i) Using the POD modes represented on a mesh grid to interpolate the modes values at $\mathbf{x}(t_i)$, (ii) solving \eqref{eq-SenLoc-approximation} for coefficients $a_k(t_i)$, and (iii) calculating the approximate measurements at $\mathbf{x}(t_i)$ using the approximated POD modes. The error is then accumulated over the entire trajectory as
\begin{equation} \label{eq-traj-err2}
    E = \sum_{i=0}^{T} \ \mathbf{e}(t_i).
\end{equation}

Another important metric to minimize is the energy (fuel) cost for the robot to navigate through the flow field along the trajectory. The relative velocity between the robot and the background flow in the deployed environment provides the kinetic energy required for the robot to move, which is accumulated over the trajectory to arrive at a net energy cost term,
\begin{equation} \label{eq-traj-enrgy}
    F = \sum_{i=0}^{T} \left[ \boldsymbol{\nu}(t_i)-\mathbf{u}^*(\mathbf{x}(t_i), t_i) \right]^2, 
\end{equation}
where $\boldsymbol{\nu}(t_i)$ is the velocity of the vehicle at time $t_i$. 

The overall cost function for trajectory optimization is given by
\begin{equation} \label{eq-traj-cost-funcn}
    J = \alpha_1 D + \alpha_2 E + \alpha_3 F, 
\end{equation}
where $\alpha_1, \alpha_2$, and $\alpha_3$ are tunable weighting constants for each of the cost terms and $D$ is the total time taken (duration) for the trajectory, i.e., $D = t_f$. The optimization problem can then be formulated as
 \begin{align} \label{eq-trajopt}
     \mathbf{z}^* = \underset{\mathbf{z}}{\mathrm{argmin}} \ J \quad
     \textit{s.t.} \quad 
     \begin{bmatrix}
         \mathbf{z}(t^{(k)}_\text{end}) = \mathbf{y}^*_k, \ k \in {1,2, \ldots , m} \\
         \mathbf{z}(t_0) = \mathbf{g}_1 \\
         \mathbf{z}(t_f) = \mathbf{g}_2 \\
        |\boldsymbol{\nu}| \leq  \nu_\text{max} \\
         t_i < t_{i+1}\\
         \mathbf{x}_\text{min} \leq \mathbf{x} \leq \mathbf{x}_\text{max} \\
     \end{bmatrix} \;.
 \end{align} 

%%%----------------- Figure:  Trajectory Illustration 

%%%----------------- Algorithm for the proposed approach
\begin{algorithm}[t!]
	\caption{Dynamic Compressed Sensing of Flow Fields}
    \begin{algorithmic}[1]
		\Require 
		Flow data ($X \in \mathbb{R}^{n \times t}$), waypoint count ($m$), number of waypoint trials ($c_1$), number of trajectory trials ($c_2$), starting location ($\mathbf{g}_1$), goal location ($\mathbf{g}_2$), and maximum vehicle speed ($\nu_\text{max}$)
		
		\Ensure Optimal sensing location set ($\mathbf{y}^*$) and optimal trajectory ($\mathbf{z}^*$)
		
		\medskip
		\Statex \textit{/* Calculate POD modes */}
		\State Calculate covariance of flow data \Comment{\eqref{eq-pod-cov}}
		\State Eigen-decomposition of covariance matrix \Comment{\eqref{eq-pod-cov-eigen}}
		\State POD mode for $i$-th location, $\phi_i = \lambda_i\mathbf{v_i}$
		
		\medskip
		\Statex \textit{/* Selection of optimal waypoints */}
		\State Calculate SVD of $X$ to obtain transform basis $U$ \Comment{\eqref{eq-svd-def}}
		\For {$i$ in $c_1$ waypoint set trials} 
		    \State Generate random measurement matrix $C \in \mathbb{R}^{m\times n}$ 
		    \State Calculate the optimal sparse vector ($\hat{\mathbf{s}}$) \Comment{\eqref{eq-comp-sens-opt2}}
		    \State Sort $\hat{\mathbf{s}}$ and pick $m$ sensing locations ($\mathbf{y}_i$)
		    \State Calculate reconstruction error ($e_i$) \Comment{\eqref{eq-SenLoc-approximation}, \eqref{eq-SenLoc-RecErr}} 
		\EndFor
		\State Get $\{\mathbf{y}^* | e_{\mathbf{y}^*} = \min (e_i) \; \forall \ i \in \{ 1, 2 \ldots c_1 \} \}$  

		\medskip
		\Statex \textit{/* Trajectory optimization */}
		\For{$j$ in $c_2$ trajectory trials} 
		    \State Randomize the waypoint sequence, $\mathbf{y}^* \leadsto \mathbf{y}^*_j$
		    \State Add start and goal locations, $\mathbf{y}^*_j = [\mathbf{g}_1; \ \mathbf{y}^*_j; \ \mathbf{g}_2]$
		    \State Define optimization cost function ($J_j$) \Comment{\eqref{eq-traj-err1} - \eqref{eq-traj-cost-funcn}}
		    \State Solve optimization problem for $\mathbf{z}^*_j$ \Comment{\eqref{eq-trajopt}}
		\EndFor
 
		\State Get $\{\mathbf{z}^* | e_{\mathbf{z}^*} = \min (J_j(\mathbf{z}^*_j)) \; \forall \ j \in \{ 1, 2 \ldots c_2 \} \}$  
	\end{algorithmic}
	\label{alg1}
\end{algorithm}

%%%%%%%%%%%%%%%%%%%%%%%%%%%%%    SIMULATION RESULTS     %%%%%%%%%%%%%%%%%%%%%%%%%%%%%%%%%%%%
\section{SIMULATION RESULTS} \label{sect:simulation}
This section presents the simulation results for application of the proposed approach on a periodic double gyre flow environment, which is inherently unsteady, making the optimization highly non-linear and challenging. The results present the details and the performance of the compressed sensing algorithm for selection of optimal waypoints, generation and implementation of an optimal trajectory through those waypoints and a comparison between the performance of the system with the number of prescribed waypoints as the variable and the components of the optimization function costs (reconstruction error, energy and total time taken) as the metric.

%%%%%%%%%%%%%%====================  Implementation on  Double Gyre
\subsection{Implementation on Unsteady double gyre Flow}
The proposed algorithms for generation of optimal waypoints and an optimal trajectory through those to observe the flow field for effective reconstruction of the entire environment were implemented on a periodic double gyre flow field. The double gyre is an unsteady flow field well studied in the fluid community for chaotic mixing and material transport. It resembles real-world flow phenomena often found in atmospheric and oceanic flow environments and bears an analytical expression, making it an ideal benchmark environment to test our algorithms. 

The double gyre flow field is defined by the stream function,
%%%%%%%%%%%%
\begin{equation}
    \psi= A \sin{(\pi f(x,t))}\sin{(\pi y)},
\end{equation}
over a non-dimensionalized domain of $[0,2] \times [0,1]$, where
\begin{equation} 
\begin{split}
    f(x,t) & = a(t)x^2+b(t)x, \\
    a(t) & =\epsilon \sin{(\omega t)}, \\
    b(t) & = 1-2\epsilon \sin{(\omega t)},
\end{split}
\end{equation}
The velocity field is given by
\begin{equation} 
\begin{split}
    u &= -\frac{d \psi}{dy}  = - \pi A \sin{(\pi f(x))} \cos{(\pi y)}, \\
    v &= \frac{d \psi}{dx} \  \ = \pi A \cos{(\pi f(x))} \sin{(\pi y)} \frac{df}{dx},
\end{split}
\end{equation}
where $A$ determines the magnitude of the velocity vectors, $\frac{\omega}{2 \pi}$ is the frequency of oscillation, and $\epsilon$ is approximately the magnitude of the periodic oscillations, as discussed in Shadden et. al.~\cite{shadden2005definition}. The value of $\epsilon$ determines the time dependency of the flow: for $\epsilon = 0$ the flow is time-independent, and for $\epsilon \neq 0$ the gyres expand and contract periodically in the x-direction conversely with the rectangle enclosing the gyres remaining invariant. 

%%%----------------- Figure:  Sensor Locations vs Error 
\begin{figure}[t]
    \centering
    \includegraphics[width=0.95\linewidth]{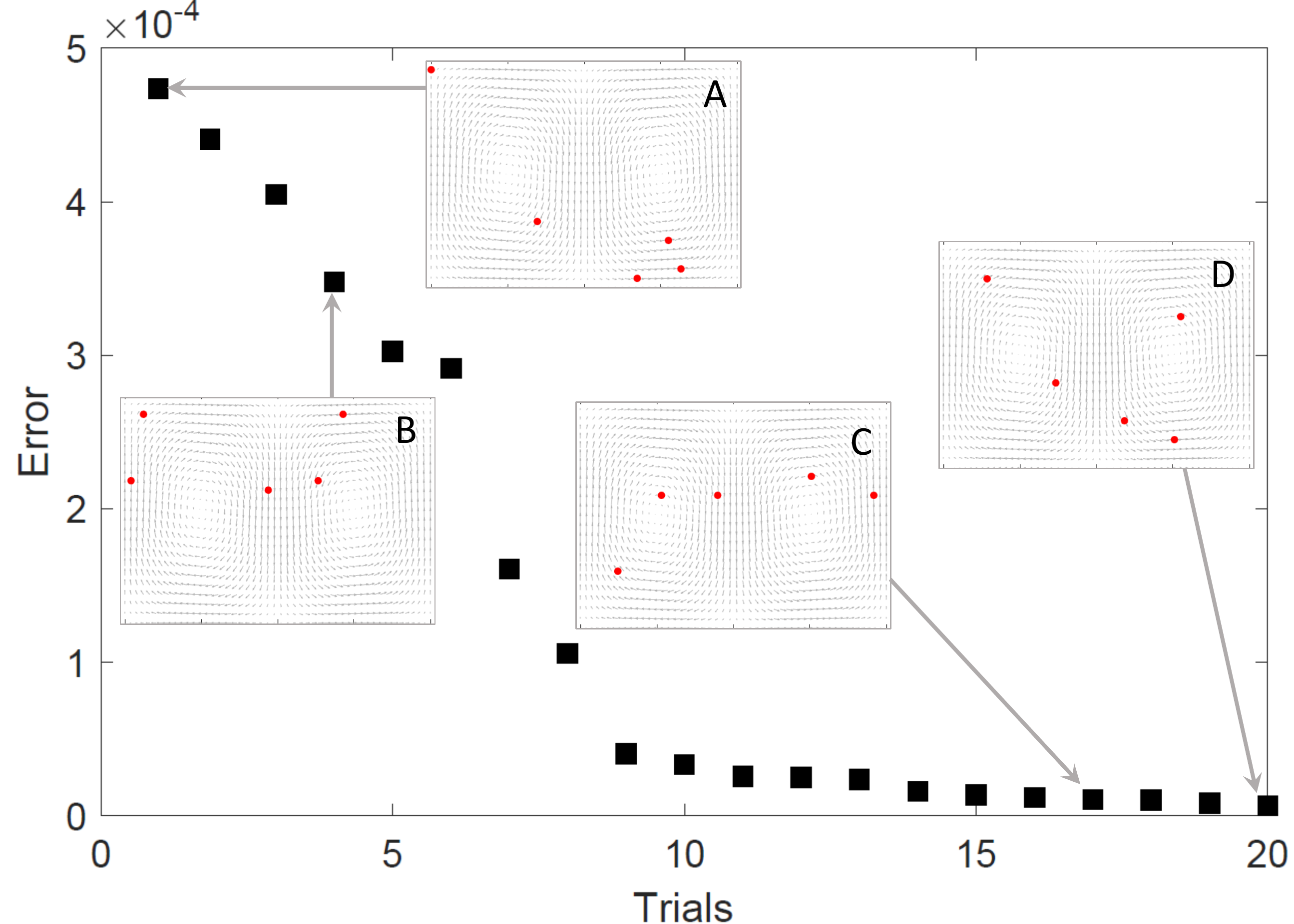}
    \caption{Reconstruction errors for the compressed sensing based unique grid-waypoint sets (five waypoints). The plot shows a total of 20 trials and their respective errors. The sub-images show four of the waypoint set options, of which set A produces the highest reconstruction error, set D produces the least and sets B and C have intermediate values for the error. Set D is chosen as the optimal waypoint sequence for trajectory optimization in the next step. The errors are calculated for the reconstruction of a periodic double gyre flow with $T=2001$, but the background vector field in each sub-image shows velocity vectors at $t=0$ only.}
    \label{fig:SenVsErr}
\end{figure}

%%%%%%%%%%%%%%====================  Compressed Sensing vs Random locations
\subsection{Compressed Sensing and Trajectory Optimization}
The first component of the proposed approach is to choose a highly reduced subset of sensing locations for effective reconstruction of a large area of interest, using compressed sensing. The reconstruction error is defined in terms of the POD modes of the flow environment, which can be generated using the definition framework of the double gyre flow field with $n_x = 50 \text{ and } n_y = 25$ (1250 grid locations total, and $x_\text{lim} = 2, \ y_\text{lim} = 1, \ t_\text{lim} = 20 \text{ with } \delta t = 0.01$). This has been done in two steps for effective selection of the waypoints at the grid locations: (i) generate a sparse representation for a randomized measurement vector for the given number of waypoints and picking the grid location subset with best reconstruction performance, and (ii) repeat this to generate twenty of such waypoint sets and choose the one with least error. This exploits the fact that the selection of locations is randomized and not a unique global solution. An example for five waypoints has been shown in Fig.~\ref{fig:SenVsErr}, where the best of twenty waypoint set is selected as the optimal subset for the following step, the optimal trajectory generation. It is to be noted that each of these waypoint sets are near-optimal themselves and have high reconstruction efficiency (very small reconstruction error). Four sample waypoint sets (A, B, C and D) have been shown in the sub-images to show the non-uniqueness of the reconstruction performance, and the one with the best reconstruction performance of the lot is chosen as the optimal solution.

\begin{figure}[t]
    \centering
    \includegraphics[width=0.95\linewidth]{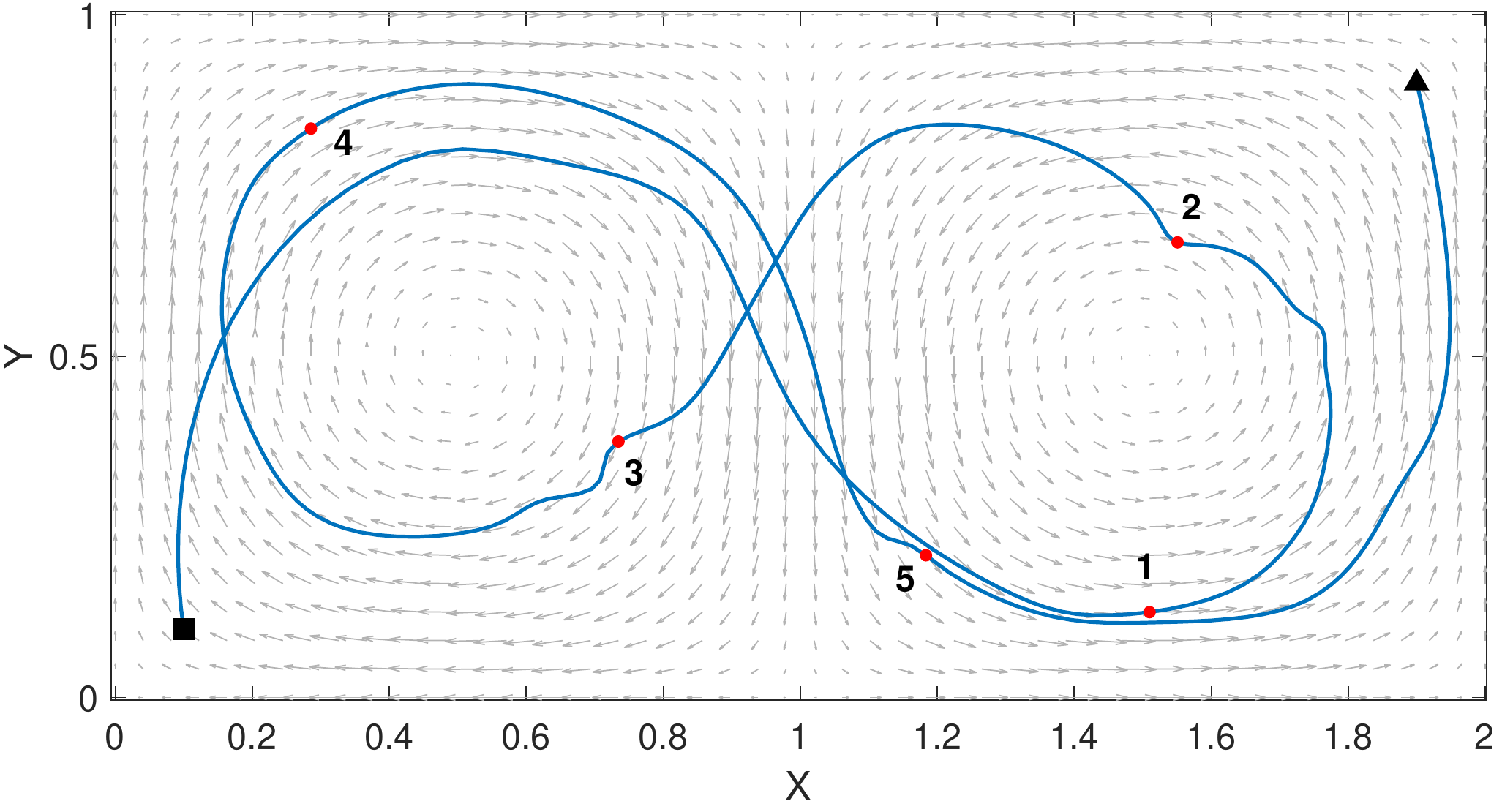}
    \caption{Dynamic compressed sensing in an unsteady physical flow: The mobile robot uses an optimal trajectory to navigate through the compressed sensing based waypoints in the periodic double gyre flow environment to collect data for reconstruction, while minimizing the reconstruction error, energy cost and total time taken. The black square, black triangle and the sequence of red dots denote the starting and end location, and the chosen waypoints, respectively. The trajectory is optimized for the reconstruction of a periodic doublegyre flow with $T=2001$, but the background vector field shows the velocities at $t=0$ only. All quantities involved in this simulation are non-dimensionalized.}
    \label{fig:DGimplementaion}
\end{figure}

\begin{figure*}[t]
    \centering
    \includegraphics[width=0.325\linewidth]{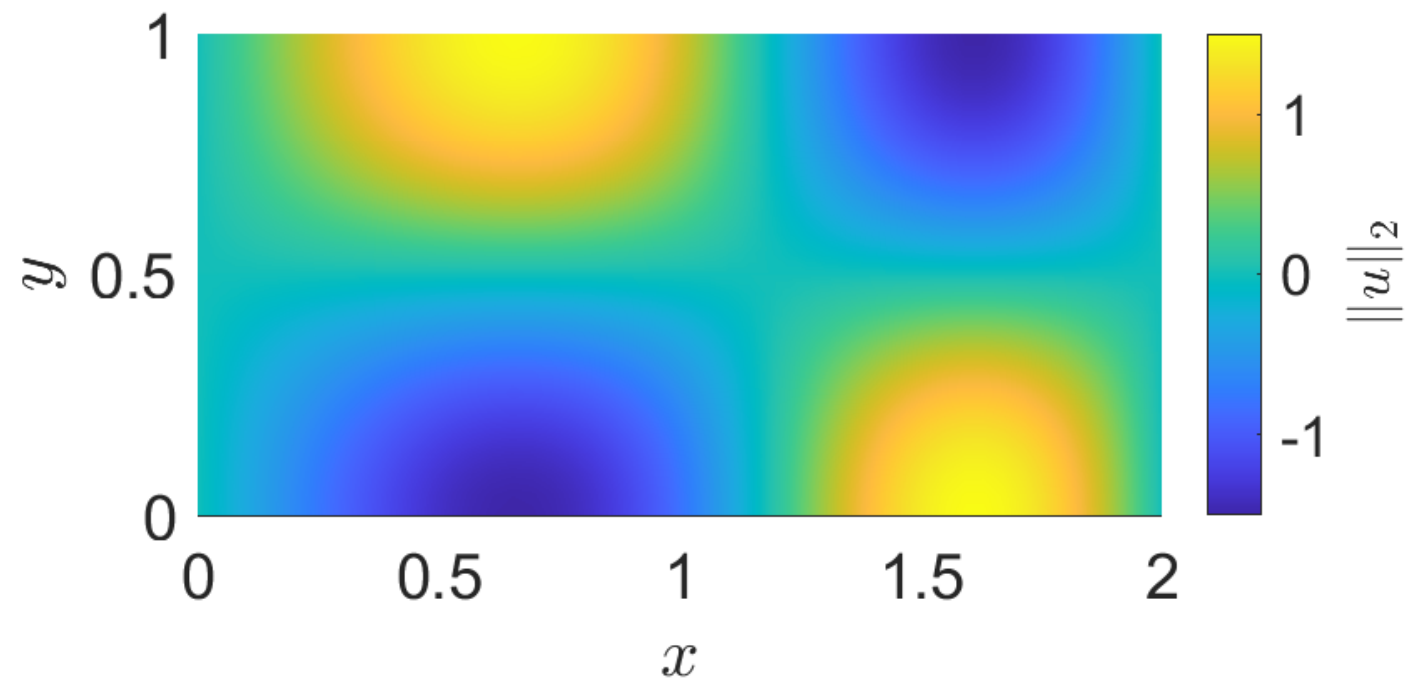} 
    \includegraphics[width=0.325\linewidth]{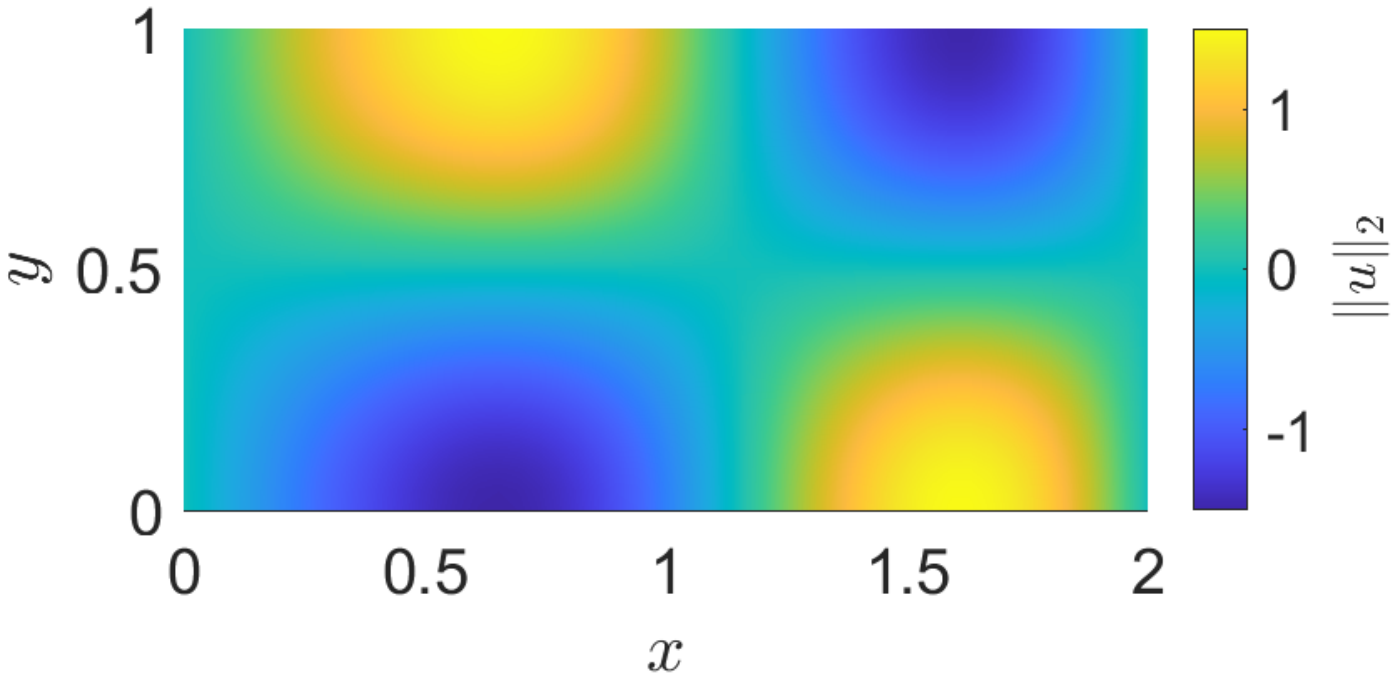} 
    \includegraphics[width=0.325\linewidth]{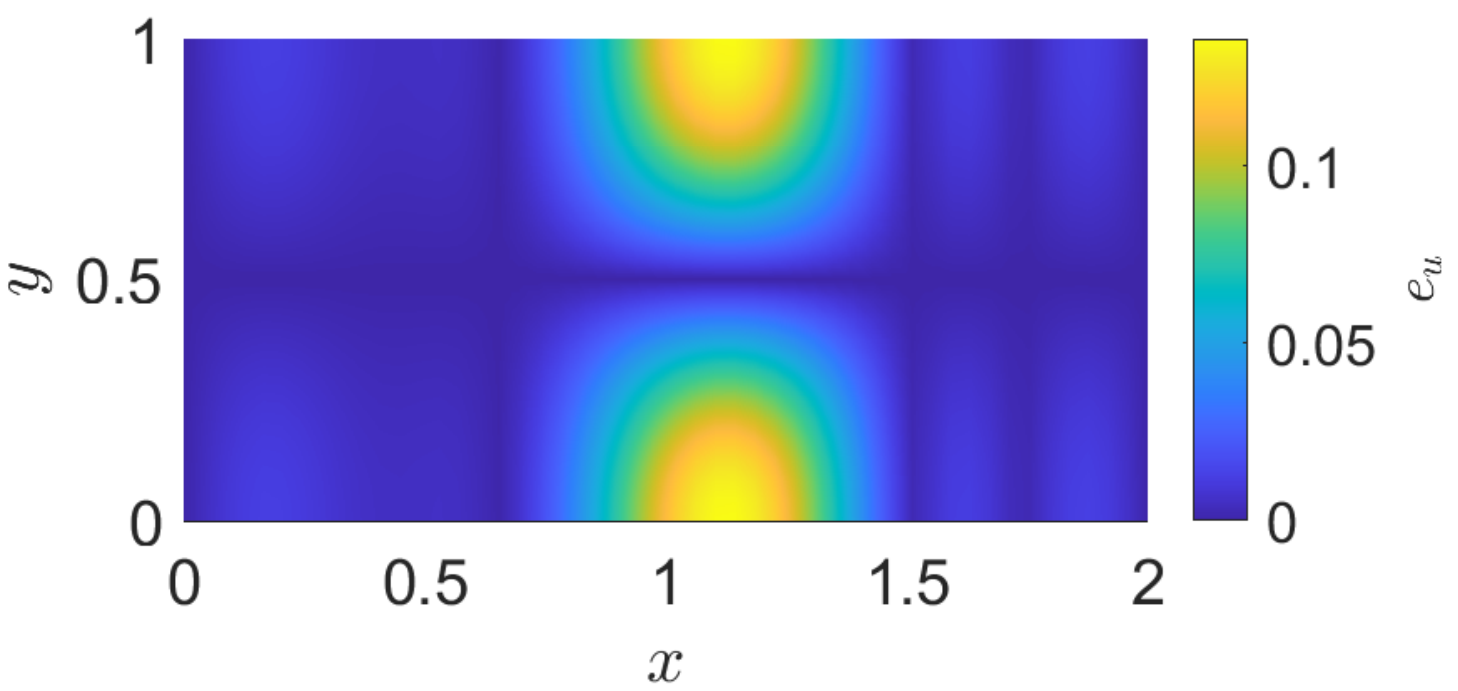} \\ \footnotesize{(a)} \hspace{5.5cm} \footnotesize{(b)} \hspace{5.5cm} \footnotesize{(c)}
    \caption{Reconstruction performance through the trajectory measurements: (a) Reference double gyre flow map ($u$-component) at $t=251$, (b) reconstructed flow at the same instant, and (c) difference between the reference and reconstructed flow. The error is mostly large at this instant as there are not enough measurements for efficient reconstruction.}
    \label{fig:Reconstruction}
\end{figure*}

In the next step, an optimal trajectory is generated through the chosen optimal waypoints (sub-image D in Fig.~\ref{fig:SenVsErr}) with a pre-specified starting and goal location as shown in Fig.~\ref{fig:DGimplementaion}, with an objective to minimize the reconstruction error, the energy cost and the total time taken. The trajectory optimization is based on generating a continuous fourth order spline-based trajectory, and solved using the \textit{fmincon} function in MATLAB, with $v_\text{min} = 0, \ v_\text{max} = 0.7,  \text{ and } \Delta t = 0.1$. We also compute and compare the trajectories between the various shuffled sequences of the chosen optimal waypoint set. The order of the waypoint sequence from the starting to the goal location has been marked in Fig.~\ref{fig:DGimplementaion}. The mathematical formulation of the selection of optimal sequence to visit a chosen set of waypoints is beyond the scope of this work. It can be seen from the figure that the figure that the generated trajectory is successfully optimizing the energy cost by adapting to the background flow field, and yields low values of accumulated costs (reconstruction error = 11.56 , energy cost = 110.3 and time cost = 6.07). The reconstruction error and the energy cost values have not been normalized for the overall grid. The reconstruction example for an instant with the first measurement ($t=1$) of the above mentioned trajectory through five waypoints has been shown in Fig~\ref{fig:Reconstruction}, where (a), (b) and (c) show the original reference double gyre flow map, measurement based reconstruction and the difference between (a) and (b), respectively. The proposed dynamic compressed sensing approach successfully chooses a set of optimal sensing locations over the desired area and then generates an energy optimal trajectory to visit them in quick time for efficient reconstruction. 

%%%----------------- Figure:  Implementation on DG 

%%%----------------- Figure:  Reconstruction performance

%%%%%%%%%%%%%%====================  Performance Evaluation
\subsection{Performance Evaluation}
\begin{figure*}[t]
    \centering
    \includegraphics[width=0.31\linewidth]{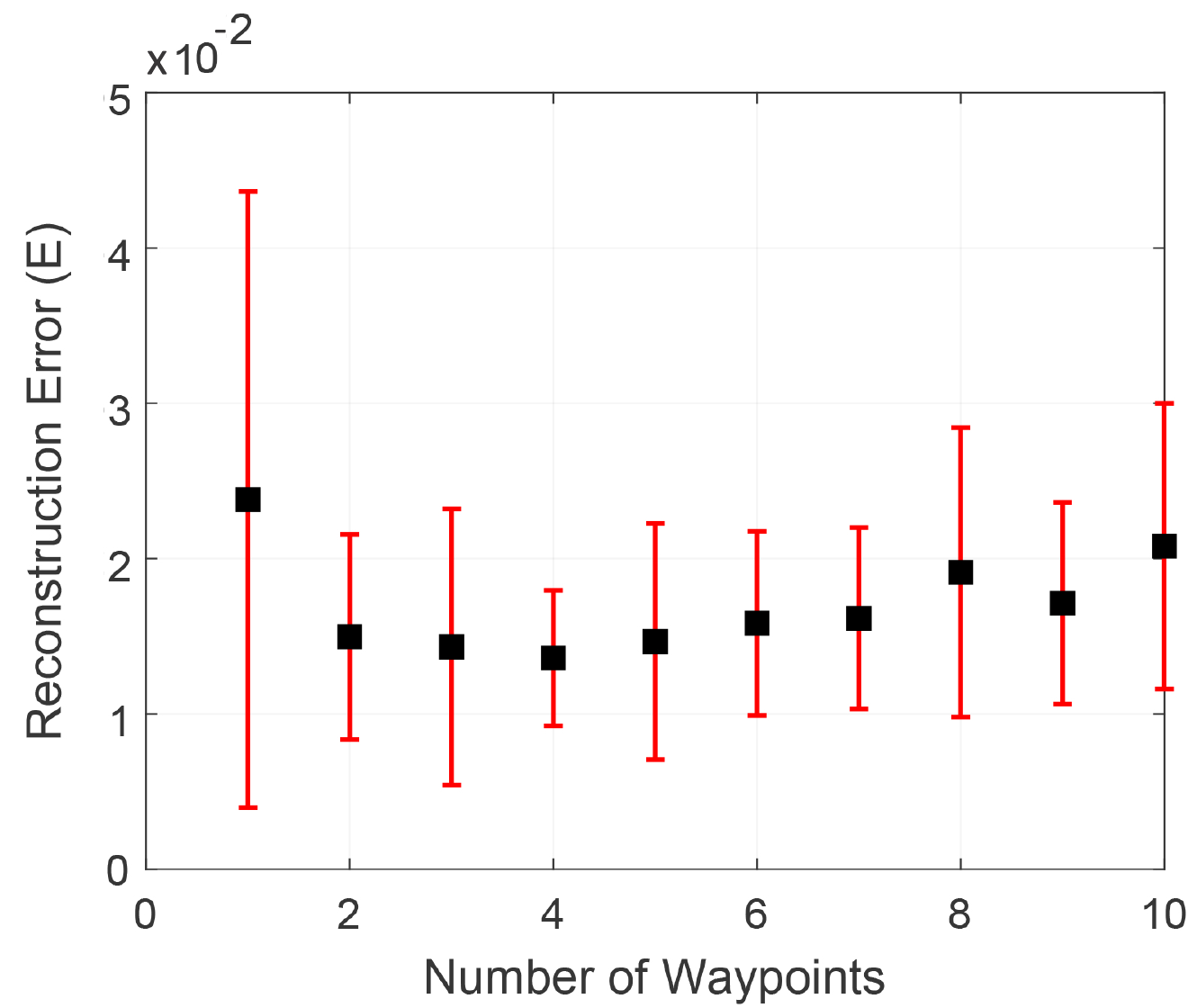} \hspace{3mm}
    \includegraphics[width=0.32\linewidth]{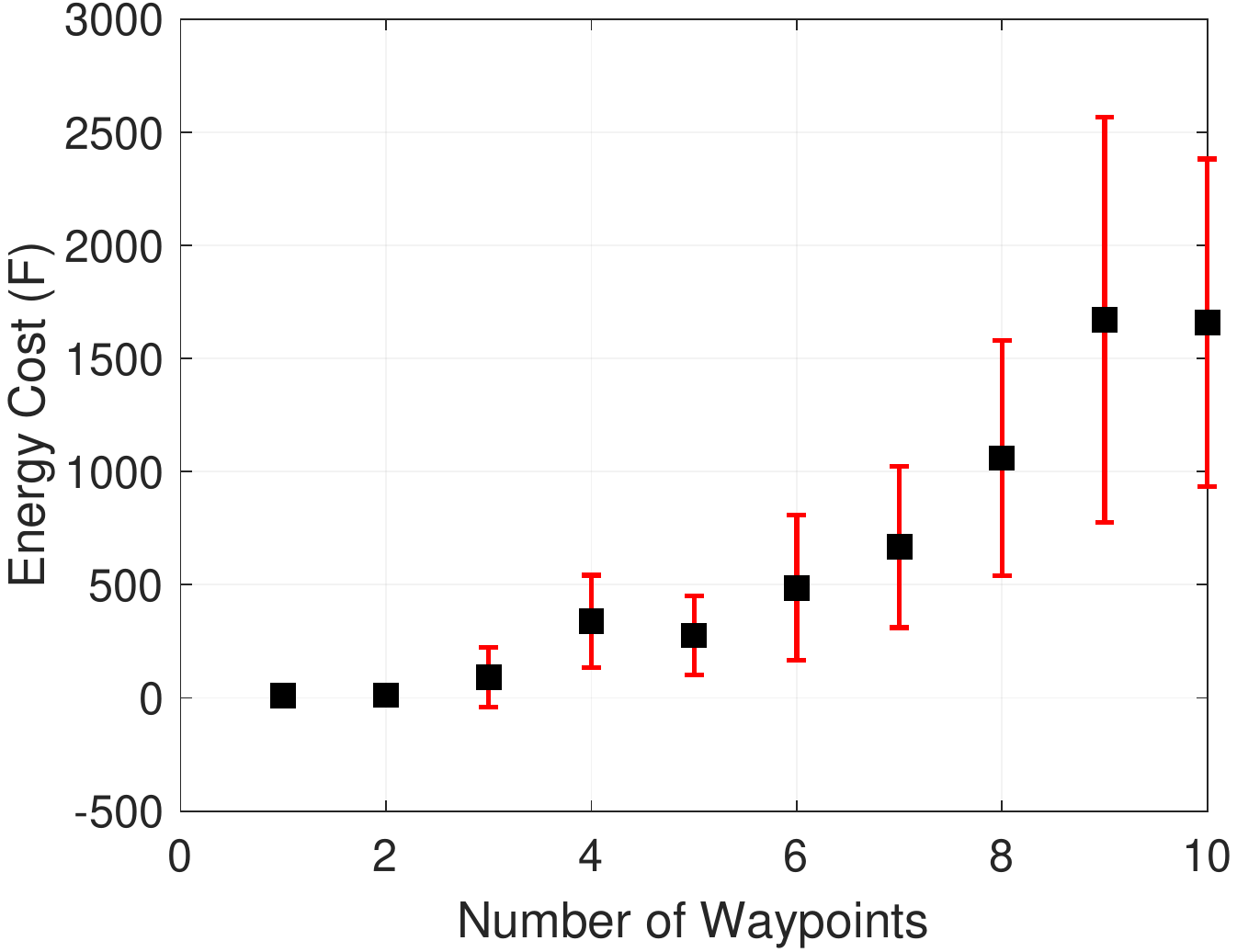} \hspace{3mm}
    \includegraphics[width=0.31\linewidth]{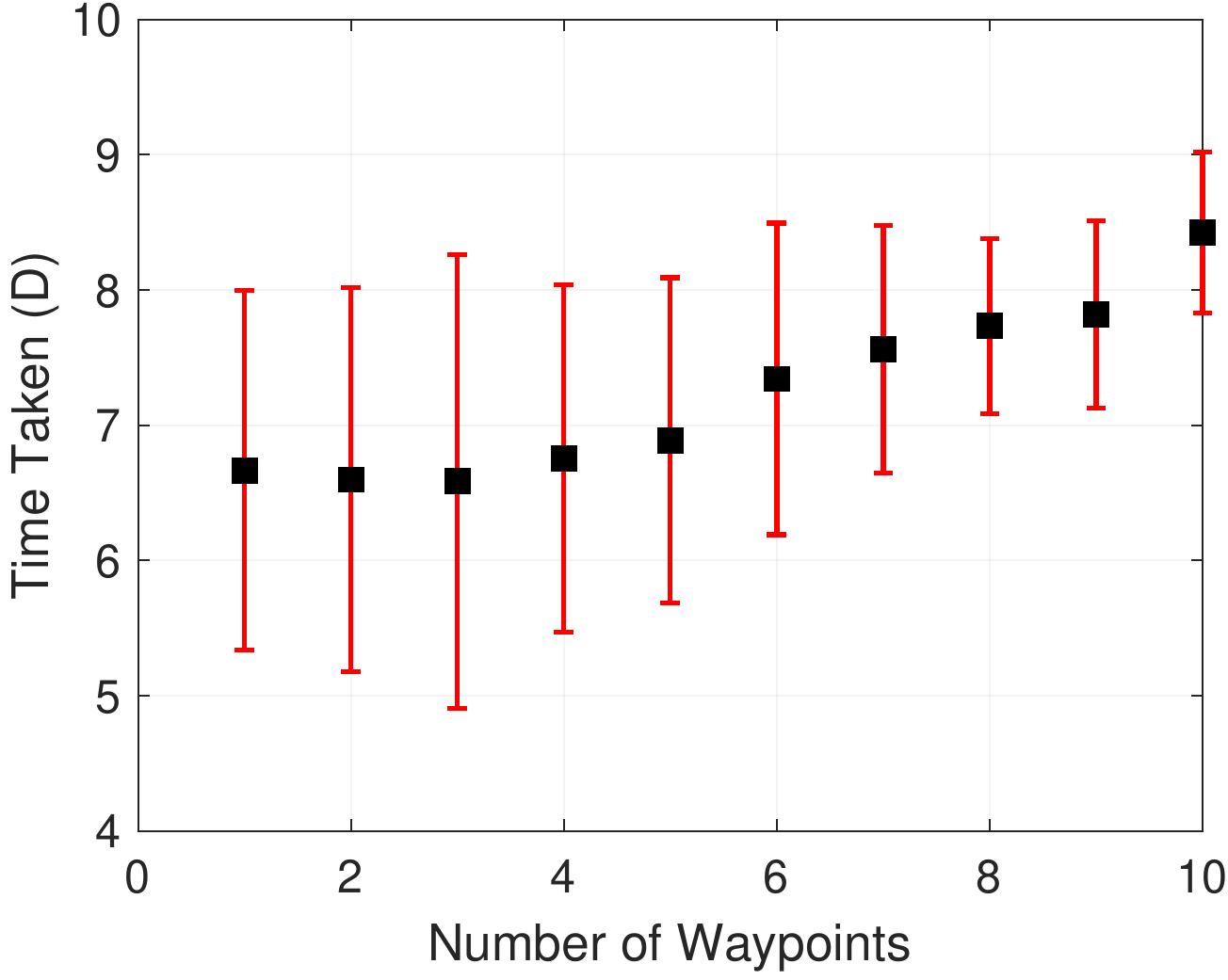} \\ \footnotesize{(a)} \hspace{5.5cm} \footnotesize{(b)} \hspace{5.5cm} \footnotesize{(c)}
    \caption{Performance evaluation of the proposed trajectory optimization algorithm for waypoint set sizes ($1 \sim 10 $): (a) Accumulated reconstruction error, (b) accumulated energy cost, and (c) total time taken for the trajectory. The plots show the average and 1-$\sigma$ values over 70 test trajectory cases for each waypoint size (10 unique set of waypoints $\times$ 7 shuffled sequence for each set). }
    \label{fig:TrajPerformance}
\end{figure*}

The proposed algorithm was implemented on a range of waypoint set sizes ($1 \sim 10 $) to check its robustness and compare performance as the waypoint size changes. For every waypoint size, ten unique sets of waypoints were generated and each of those was shuffled seven times to generate various trajectories through same set of waypoints. This resulted in a total of 70 unique trajectories, and the performance metric (Reconstruction error, Energy cost and Total time) was averaged, which has been shown in Fig.~\ref{fig:TrajPerformance} along with the $1- \sigma$ values for each metric.

The average accumulated reconstruction error performance for each of the waypoint size has been shown in Fig.~\ref{fig:TrajPerformance}(a). As expected, the reconstruction is performed very effectively, and is comparatively worse with only one waypoint as the measurements for those cases would not be enough for effective reconstruction. The performance improves with increasing number of waypoints but quickly saturates as the number of measurements over the trajectories become enough for the effective reconstruction, and adding more waypoints does not contribute any additional information value. But overall, these are satisfactory results and in sync with the energy cost performance shown in Fig.~\ref{fig:TrajPerformance}(b). The algorithm does not need to force the equality constraints to meet the control requirements for lesser number of waypoints, and the trajectory is mostly driven by the background flow. As the number of waypoints increases, the control effort from the algorithm increases for the trajectories to visit the waypoints. Combined with the relatively larger length of the trajectories (which adds to the possible travelling against the flow), the overall energy cost is increased. Finally, the time taken for a trajectory though a given number of waypoints under the influence of a background flow should intuitively increase with the increase in number of waypoints, as the trajectory would become longer to optimize energy. The results for total travel time in Fig.~\ref{fig:TrajPerformance}(c) are mostly consistent with this intuition, with some inconsistencies through the average and the standard deviation values. The inconsistencies and unexpected results can be attributed to the randomness of the data and parameter sensitivity of the optimization algorithm to some extent, but this leaves the scope for further improvement of the proposed approach. Despite the data set being highly random, the results are consistent and as intuitively expected. This shows that the proposed approach is effective for a large range of application scenarios. 
 
%%%----------------- Figure:  Comparison between nWpts performance

%%%%%%%%%%%%%%%%%%%%%%%%%%%%%%%%%%%%    EXPERIMENTAL RESULTS     %%%%%%%%%%%%%%%%%%%%%%%%%%%%%%%%
\section{EXPERIMENTAL RESULTS} \label{sect:experiment}
This section presents experimental results for validation of the proposed approach using the Bitcraze quadcopter experimental platform (Crazyflie 2.0) and a Vicon motion capture system, where a Crazyflie emulates the simulated optimal trajectory for an unsteady flow field. The basics of the experimental setup with Crazyflies and Vicon has been discussed in~\cite{preiss2017crazyswarm}. Luis and Ny~\cite{luis2016design} also presented a method for using Crazyflie for trajectory following. In this experiment, the simulated trajectory positions were uploaded onto the Crazyflie and the platforms replicate the simulated trajectory. Using the Vicon camera system and the tracking interface, we were able to record the location and orientation of the quadcopter, and overlap that recording over the simulated trajectories. 

\begin{figure*}[t]
    \centering
    \includegraphics[width=0.31\linewidth]{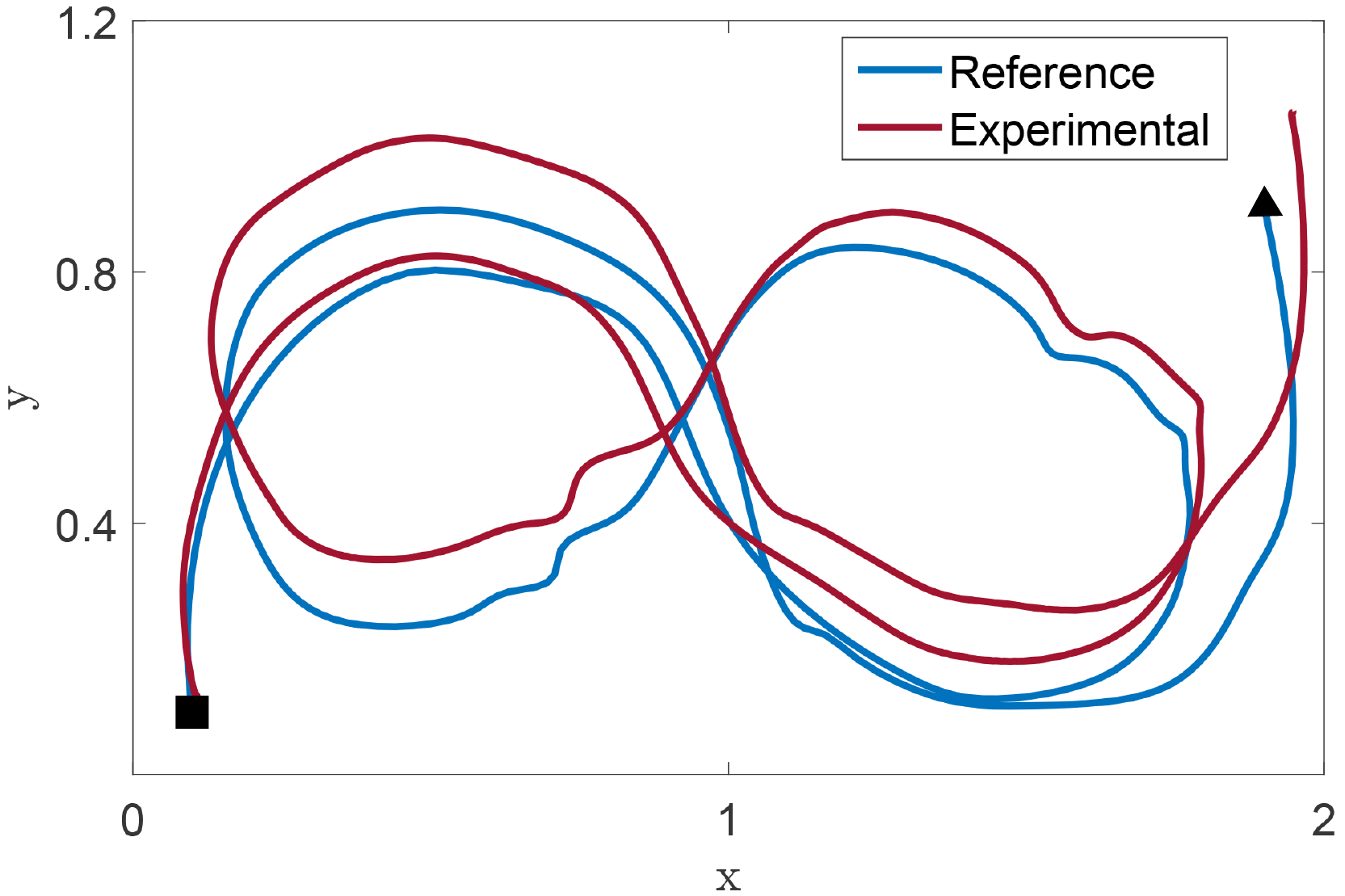} \hspace{3mm}
    \includegraphics[width=0.31\linewidth]{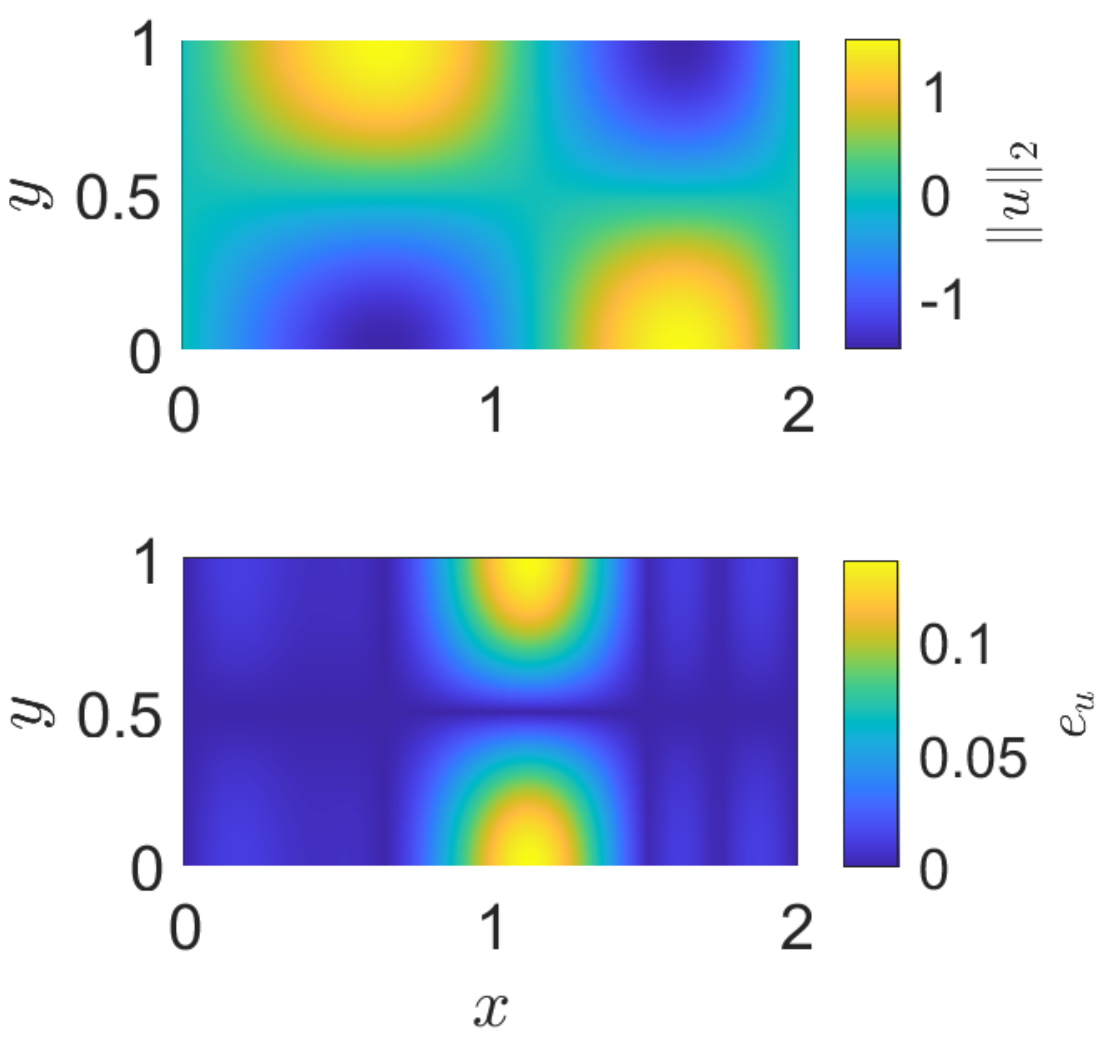} \hspace{3mm}
    \includegraphics[width=0.31\linewidth]{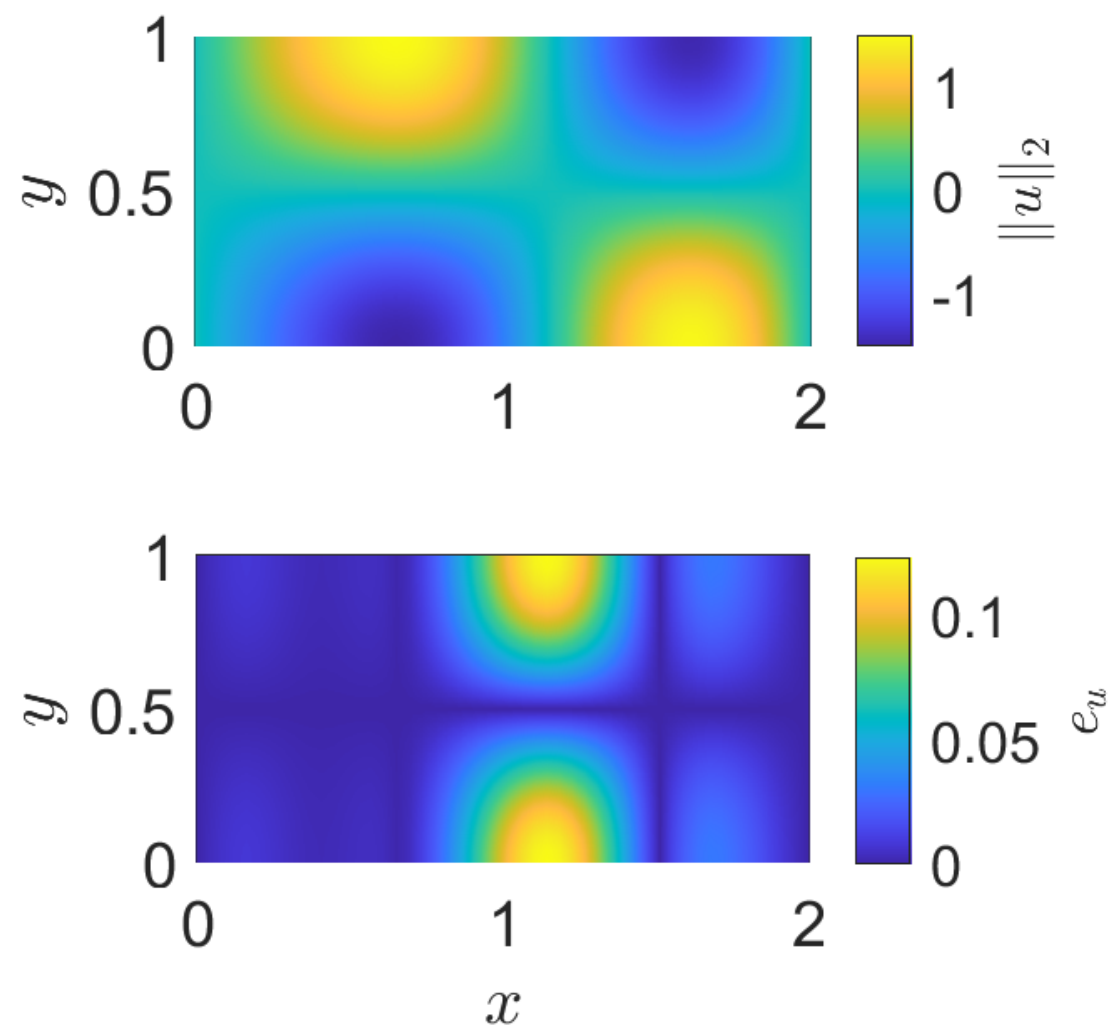} \\ \footnotesize{(a)} \hspace{5cm} \footnotesize{(b)} \hspace{5.5cm} \footnotesize{(c)}
    \caption{Experimental results for a Crazyflie quadcopter following a simulated reference trajectory. (a) Top view of the flight path for the quadcopter while following the simulated reference trajectory, (b) reconstructed $u$-velocity map (top) and error map (bottom) compared to the ideal double gyre at $t=251$ for the reference trajectory, and (c) equivalent reconstruction using the experimental trajectory.}
    \label{fig:expt}
\end{figure*}

The experimental flight path of the quadcopter with respect to the simulated trajectory can be seen in Fig.~\ref{fig:expt}(a), along with the prescribed origin (square) and destination (triangle). It can be seen that the quadcopter can effectively track the trajectory with some positional drift and offset. These offsets are caused by faulty measurements by the inertial measurement unit (IMU) of the quadcopter, or the hardware limitations due to the close proximity of the adjacent reference trajectory positions. Figure~\ref{fig:expt}(b) shows a reconstruction time instant ($t=51$ for trajectory of length 305 units) and its corresponding error (difference from the ideal flow field at that instant) for the simulated reference trajectory in  Fig.~\ref{fig:expt}(a), stacked on top and bottom, respectively. Figure~\ref{fig:expt}(c) is the equivalent for the experimental trajectory from the quadcopter flight. It can be seen that the simulated trajectory near-accurately reconstructs the underlying flow field, and the experimental trajectory closely follows with small error values, despite the offset between the trajectories. The overall results show that it is viable to implement this algorithm for inherently unsteady flow environments, and can be further improved by tuning the optimization parameters while generating the trajectory, taking into account the sensitivity of the experimental setup being used.

%%%----------------- Figure:  Experimental trajectory

%%%%%%%%%%%%%%%%%%%%%%%%%%%%%%%%%%    CONCLUSIONS     %%%%%%%%%%%%%%%%%%%%%%%%%%%%%%%%%%%%%%%%%
\section{CONCLUSIONS} \label{sect:conclusion}
This paper presents the dynamic compressed sensing approach, which aims to reduce the cost compared to the compressed sensing by deploying a mobile robot to take measurement instead of stationary sensors, based on the assumption that not all locations need to measure data at each time instant. The algorithm first chooses a reduced subset of optimal sensing locations with efficient reconstruction performance using compressed sensing and then uses those locations as waypoints to plan an optimal trajectory to effectively reconstruct the overall information, while minimizing the energy cost and total travel time for the trajectory. Simulation results show that the proposed approach effectively optimizes the sensing locations and trajectory, and works well for a range of scenarios. Furthermore, the experimental results validate the implementation of the approach. 

The proposed method is currently limited to periodic flow fields and assumes accurate knowledge of the POD basis. We are currently expanding the method to handle turbulent flow effects that are not capture by the data used to construct the POD basis. For general flow fields that do not possess periodicity, we plan to adopt a distributed dynamic compressed sensing scheme with multiple mobile sensors to better capture the spatiotemporal correlations among flow field sampling results.

%%%%%%%%%%%%%%%%%%%%%%%%%%%%%    REFERENCES    %%%%%%%%%%%%%%%%%%%%%%%%%%%%%%%%%%
\bibliographystyle{IEEEtran}
\bibliography{references}

\end{document}